\documentclass{llncs}
\usepackage{makeidx}  

\usepackage{graphicx,psfig} 
\usepackage{amsfonts}
\usepackage{amsmath} 

\newcommand{\argmin}{\operatornamewithlimits{argmin~}}

\begin{document}
\frontmatter          
\pagestyle{headings}  
\title{Object Recognition with Multi-Scale Pyramidal Pooling Networks}
\titlerunning{Object Recognition with Multi-Scale Pyramidal Pooling Networks}  
%
\author{Jonathan Masci\inst{1} \and Ueli Meier\inst{1} \and
Gabriel Fricout\inst{2} \and J\"{u}rgen Schmidhuber\inst{1}}
\authorrunning{Masci et al.} 
%
\tocauthor{Jonathan Masci, Ueli Meier, Gabriel Fricout, J\"{u}rgen Schmidhuber}
\institute{IDSIA -- USI -- SUPSI, Manno -- Lugano, Switzerland,\\
\email{jonathan@idsia.ch}, \texttt{http://idsia.ch/\homedir masci/}
\and
ArcelorMittal, Maizi\`{e}res Research, Measurement and Control Dept., France}

\maketitle              

\begin{abstract}

We present a Multi-Scale Pyramidal Pooling Network, featuring a
novel pyramidal pooling layer at multiple scales and a novel
encoding layer. Thanks to the former the network does
not require all images of a given classification task to be of equal size. The
encoding layer improves generalisation performance in comparison to similar
neural network architectures, especially when training data is scarce. We
evaluate and compare our system to convolutional neural networks
and state-of-the-art computer vision methods  on various benchmark datasets. 
We also present results on industrial steel defect classification, where existing architectures are not applicable
because of the constraint on equally sized input images. The proposed
architecture can be seen as a fully supervised hierarchical bag-of-features extension that is
trained online and can be fine-tuned for any given task.

\keywords{neural networks, pattern recognition, industrial application, bag-of-features, computer vision}
\end{abstract}

\section{Introduction}
Localising and recognising generic objects within varying scenes is
of crucial importance for computer vision and machine learning applications.
The difficulty of this task is due to high view point-dependent variability and  high in-class variability of certain
objects. Nevertheless, recognition algorithms are already quite successful in
recognising and localising objects. This success can be attributed to orientation
and scale invariant feature extractors \cite{lowe:1999,dalal:2005,Bosch:2007,Bay:2008}  or to dedicated neural network architectures,
able to capture the two-dimensional structure of images and
learn invariant representations \cite{lecun-bengio-94,lecun-bengio-95a,simard-99,lee:2009}.

In computer vision one winning strategy is based on the so-called bag-of-features approach \cite{sivic:2003,csurka:2004}. For a given image a set of features, such as SIFT descriptors, are extracted and then encoded in an overcomplete sparse representation using a dictionary-based coding technique. This produces a histogram of ``active'' words representative of the content of the image. The feature extraction is hard-coded whereas the dictionary is specifically built for a given task. That is, for classification tasks feature vectors from all or a subset of images from the training set are collected and clustered. The cluster centres form the basis of the dictionary that is used in the coding stage. Finally a supervised classifier is trained to classify the histogram representation of the image. 

Machine learning methods, on the other hand, try to map the pixel-based representation directly into a label vector, learning the feature extraction and coding from a labelled dataset. 
%
Among the most successful methods for image classification are variants of Convolutional Neural Networks \cite{fukushima:1980,lecun:1998,behnke:2003,ciresan:2011b} reminiscent of simple and complex cells in the primary visual cortex \cite{hubel:1968}.
Learning good features for such models, especially in cases where the number of training samples is scarce, opened up the investigation of many unsupervised algorithms which can be used as a pre-training stage.
This approach has been quite popular and is widely used \cite{hinton:2006,ranzato:2007,vincent:2008,coates:2010,masci:2011} to obtain better feature extractors in such setups. It is very effective for clustering or dimensionality reduction \cite{hadsell-chopra-lecun-06} but whether such pre-training leads to improved recognition accuracy for classification tasks is questionable though \cite{Rigamonti:2011}. As long as a labelled dataset is available, a fully supervised approach seems to be favourable and is the winning strategy in many benchmarks \cite{Ciresan:2012b}. A main issue remains how to effectively avoid over-fitting and improve generalisation capabilities to previously unseen images, which in many cases might be very different from the training images.  For handwritten characters elastic distortions and deformations are the best way to avoid over-fitting and improve generalisation properties \cite{simard:2003,Ciresan:2011c}. The same approach has successfully been applied to Chinese characters \cite{icdar2011Chinese}. For general object recognition tasks, the desired invariances are not that easily synthesised. Using affine transformations such as scaling, translation and rotations is the key to avoid over-fitting \cite{ciresan:2011a,2011arXiv1102.0183C,ciresan:2011b}. 

When it comes to more difficult tasks with the same object at different scales and varying location within the image, as in almost all problems of interest for computer vision, the bag-of-features approach often excels when histograms are produced at different levels \cite{Lazebnik:2006}. Pyramidal pooling is nowadays the gold standard; recent improvements are rather due to new coding strategies. Most notably linear local coding (LLC) \cite{wang:2010} led to improved results on various object recognition benchmarks.

There are obvious similarities between the bag-of-features approach and the fully supervised CNN. Both extract features based on photometric discontinuities (e.g. edges) either engineered or learnt from samples followed by an encoding stage. CNN lack of the multiple resolution pooling and moreover of an explicit encoding step which mimics the winning approaches of the bag-of-features methods. The latter lacks in a tuneable feature extraction stage which can leverage the need of a complex encoding step. Additionally CNN have the main drawback in being constrained to have constant input sized images, which can be overcome by resizing and padding, but still quite limiting in many applications.

In this paper we further close the gap between these two approaches and introduce an extension of commonly used CNN, consisting of three new ingredients: 1) the Pyramidal Pooling layer which produces a fixed-dimensional feature vector independent of the input image size; 2) a coding layer to incorporate commonly used coding strategies and 3) a multi-scale feature extraction to capture regularities visible  only at certain scales.


\section{Related Works}
Almost all object recognition algorithms use the following scheme: feature extraction, feature encoding, classification. It is desirable to obtain a linearly separable code that can be effectively classified with a linear SVM for example. For real-world applications with a very large amount of data and many different classes \cite{pascal-voc-2007,imagenet_cvpr09} this is as a matter of fact paramount, since non-linear SVM classifiers are computationally just too expensive.
In what follows we briefly recall the two main concepts of a bag-of-features system which will be reinterpreted in our Multi-Scale Pyramidal Pooling (MSPyrPool) architecture.

\subsection{Feature Encoding Algorithms}
\label{sec:featencoding}
In the computer vision framework, features are extracted using engineered approaches such as the widely used SIFT descriptors. What usually varies is not the feature extraction procedure per-se but where in the image to extract the descriptors. The most successful methods extract features over a dense, equally spaced grid \cite{Bosch:2007}. Recent improvements in classification performance on commonly used benchmarks \cite{wang:2010} are rather due to improved encoding strategies than new feature descriptors.
In order to produce a histogram, the descriptors need to be quantised such that they can be matched against a given codebook (set of basis). This step is crucial to produce an encoding with the right level of detail that avoids overfitting and hence leads to improved generalisation to unseen data. The de-facto standard for this procedure seems to be given by over-complete and sparse encodings of the feature descriptors.

Here we briefly review three of the most commonly used algorithms for feature encoding, ranging from the simplest up to the current state-of-the-art method and later we show how these approaches can be used in the fully supervised framework of our MSPyrPool network.

\begin{enumerate}

\item \textbf{Vector Quantisation (VQ)}. In its original formulation, the spatial pyramid matching uses k-means to obtain a dictionary of centroids (clusters), $\mathbf{B} = [b_{1}, b_{2}, \ldots, b_{N}]$, which is then used to solve the following least squares problem
\begin{eqnarray}
\label{eq:vq}
	\argmin_{\mathbf{C}} & = & \sum_{i}^{N} || x_{i} - \mathbf{B}c_{i} ||^{2} \\
		\nonumber
		     s.t. & \forall i . & ||c_{i}||_{0} = 1, ||c_{i}||_{1} = 1, c_{i} \ge 0
\end{eqnarray}
where the $\ell^{0}$ constraint ensures that only one basis will be on and the $\ell^{1}$ constraint ensures that its coefficient value will be $1$. In practice a $1$ is placed at the position of the nearest neighbour centroid, producing an approximation of $x_i$ using a single code vector.

\item \textbf{Sparse Coding (SC)}. Using just a single basis vector results in high quantisation errors. Relaxing the $\ell^{0}$ constraint in eq.~\ref{eq:vq} allowing more than one active basis reduces the quantisation error dramatically. Since the number of bases is bigger than the number of input dimensions the resulting system is underdetermined. An effective way to find a solution is to impose a sparsity regulariser, which reformulates the problem in a conventional and well studied sparse coding problem:

\begin{eqnarray}
\label{eq:sc}
	\argmin_{\mathbf{C}} & = & \sum_{i}^{N} || x_{i} - \mathbf{B}c_{i} ||^{2} + \lambda ||c_{i}||_{\ell^{1}}
\end{eqnarray}

This produces better reconstruction and together with a linear SVM outperforms non-linear approaches on the Caltech-101 benchmark \cite{yang:2009}.

\item \textbf{Locality-constrained Linear Coding (LLC)}. A more powerful quantisation algorithm \cite{wang:2010} exploits the locality principle to obtain sparse representations as ``locality leads to sparsity but not vice-versa''.
LLC solves the following equation:

\begin{eqnarray}
\label{eq:llc}
	\argmin_{\mathbf{C}} & = & \sum_{i}^{N} || x_{i} - \mathbf{B}c_{i} ||^{2} + \lambda ||d_{i} \odot c_{i}||^{2}
\end{eqnarray}
where $\odot$ denotes the element-wise multiplication and $d_i$ represents a regulariser to favour quantisation using bases similar to $x_i$.
LLC is very effective in commonly used benchmarks such as Caltech-101, Caltech-256 and PASCAL VOC and can be computed very efficiently using only the k-nearest neighbours of the dictionary to reconstruct $x_i$ instead of explicitly solving eq.~\ref{eq:llc}.

\end{enumerate}

\subsection{Feature Pooling}
Once the features are encoded a histogram is formed. The na\"ive approach, used in early bag-of-features systems, is to sum all the $N$-dimensional codes, where $N$ represents the number of bases in the dictionary, thus producing a global representation. A more powerful histogram generation technique is presented in \cite{Lazebnik:2006}, where features are considered in their spatial locality and representations at different levels are extracted using a quad-tree. At every level $l$ of a quad-tree $2^l$ tiles are produced and for each tile a feature vector is extracted.
The same approach is also used to produce PHOG~\cite{Bosch:2007} descriptors, an improvement over  HOG~\cite{dalal:2005}. In conjunction with a pyramid of features several methods to pool over the quadrant have been presented, such as sum-, average-, and $\ell^{2}$-pooling.

\section{Multi-Scale Pyramidal Pooling Network}
CNN and bag-of-features approaches share many basic building blocks. In both cases the feature extraction is performed using convolutional filters, with the only difference being that the filters of a CNN are learnt from the data whereas fixed filters (feature extractors) are used in bag-of-feature approaches. Furthermore, bag-of-features approaches are inherently single layer, whereas deep multilayer architectures consisting of many layers of feature extraction are capable of extracting more powerful features \cite{jarrett-iccv-09}.

%

If we interpret a CNN as a composition of many functions (layers), we obtain

\begin{equation}
\label{eq:cnncomp}
CNN(x) = f_{c} \ldots f_{k} \ldots f_{e} \ldots f_{1}(x)
\end{equation}

\noindent where $f_{1}$ to $f_{e}$ represent the feature extraction layers, $f_{e}$ to $f_{k}$ the encoding layers and $f_{c}$ the remaining classification layers.
Providing a differentiable definition for each layer results in a framework whose parameters can be jointly learnt from a training dataset.

In what follows we introduce two new layers that are used in our Multi-Scale Pyramidal Pooling Network.

\subsection{Pyramidal Pooling Layer}
%
Previous work \cite{NIPS2011_0538} uses a dynamic pooling layer to obtain features independent of input size, to detect paraphrases of 1D signals. Here we present a variation which takes into account the 2D nature of images and produces spatially located representations at several resolutions \cite{Lazebnik:2006}. Max-pooling further improves performance of the pooling operation.
The last layer of the feature extraction part can be constrained to extract a 
histogram-like 
representation at multiple levels in the spatial pyramids. Let us consider a layer with $k$ maps, each map corresponding to a filtered/pooled image. If all the values in each of those maps are summed, a $k$-dimensional feature vector which does not depend on the actual input image size, is obtained.
This simple idea produces a representation that only depends on the number of maps and which already represents a major improvement over conventional CNN.

However, summing all the activations of a map results in higher values for bigger images. To obtain a more stable measure, average pooling is preferred. An even more effective way of performing feature pooling uses the max operator instead of the average. This avoids normalisation all together and in our experiments always speeded up learning. Hereafter we consider only pyramidal pooling layers with max-pooling.

\begin{figure}[htbp]
\begin{center}
\includegraphics[width=.9\textwidth]{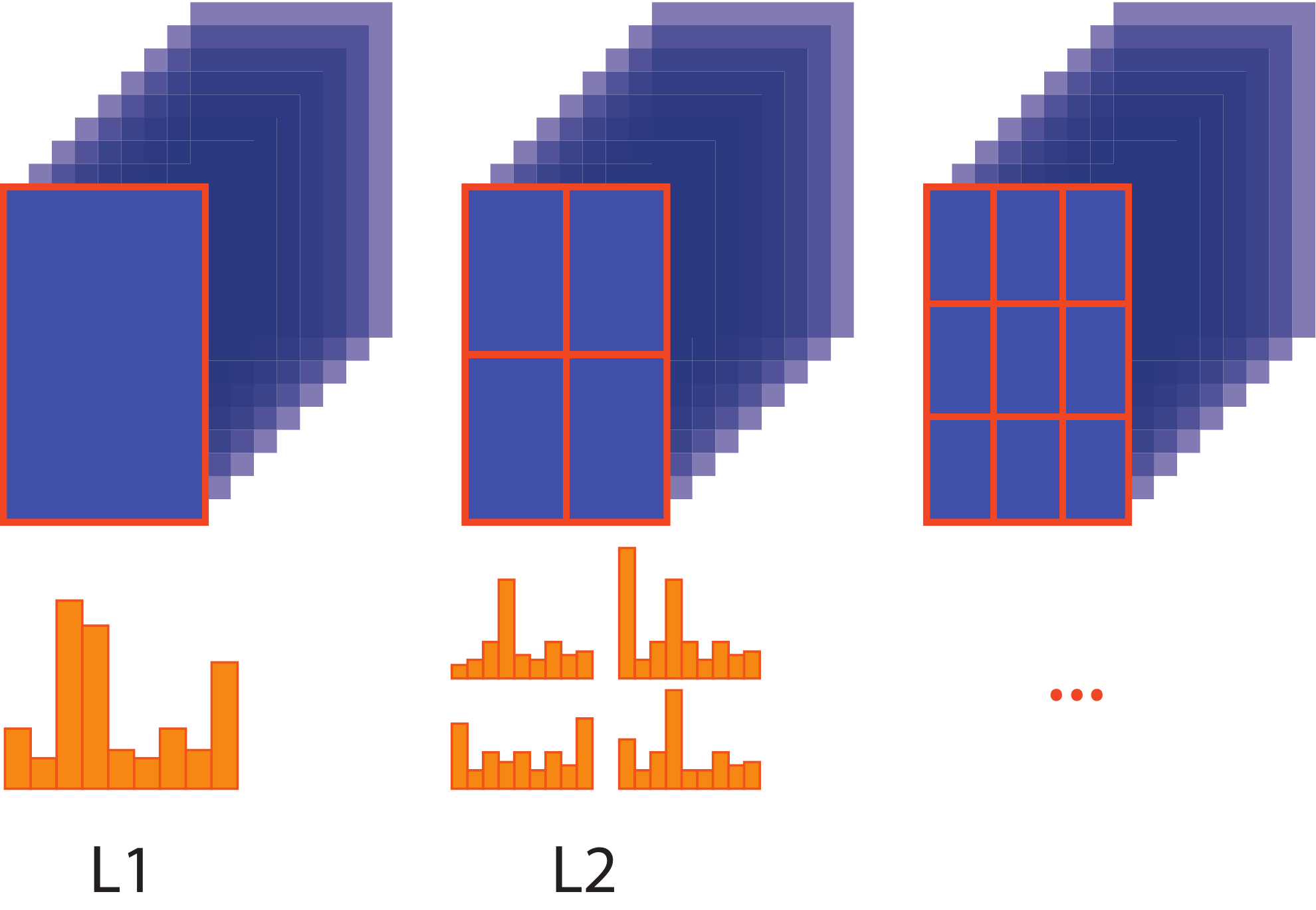}
\caption{Schematic representation of a Pyramidal Pooling Layer. The features are pooled along $l^2$ equally sized quadrants and the histogram-like representations are concatenated to form a feature vector. The pooling is preferably performed using the max operator.}
\label{fig:pyrpool}
\end{center}
\end{figure}

Still the resulting representation remains global. That is, a feature vector which is not able to capture the spatial locality of the features and their correlations within the various parts of an image. To alleviate these shortcomings the image is divided in a quad-tree fashion. A feature vector is extracted for each of the quadrants dynamically changing the pooling size according to the number of tiles which are required at each pyramid level. 
The final representation is then obtained concatenating the results of each of the levels, just as in the bag-of-feature approach. 
A graphical representation of the pyramidal pooling concept is shown in Figure \ref{fig:pyrpool} and closely follows \cite{Lazebnik:2006}. 

The Pyramidal Pooling layer does not have any tuneable parameters but in order to train the feature extraction layers before the pyramidal pooling layer the errors need to be back-propagated (that is the partial derivative of the layer's output w.r.t. its input is required).

Let us recall how the conventional max-subsampling operation works and let us denote by $\mathbf{X}$ the 3-dimensional input vector (e.g. $[\#rows, \#cols, \#maps]$). During the forward pass only a value in every non overlapping sub-region of $\mathbf{X}$ is preserved. Hence the image gets down-sampled by a constant factor given by the pooling size (e.g. usually $2\times2$). The backward pass places the delta values (results of partial differentiation by applying the chain-rule) at the location at which the maxima value was found, producing an output sized as $\mathbf{X}$. In our pyramidal pooling layer, the forward pass is equivalent to applying a subsampling operation at each level in the pyramid, built over $\mathbf{X}$, and then concatenating the result into a single output vector. Consequently the backward pass corresponds to the sum of the back-propagation of each of the subsampling operations. 

Let us denote with
\begin{equation}
\mathrm{forward:} \Pi_{l}(\mathbf{X}) \quad \quad \quad \mathrm{backward:} \frac{\partial \Pi_i}{\partial \mathbf{X}}
\end{equation}
\noindent the conventional pooling operations where pooling size is fixed at $\textrm{size}(\mathbf{X})/l$, producing $l^2$ sub-regions.

The forward pass of a Pyramidal Pooling layer, where $\textrm{cat}$ concatenates a set of vectors, can be expressed as
\begin{equation}
y = \forall i \in l \quad . \quad \textrm{cat}(\Pi_i(\mathbf{X}))
\end{equation}
\noindent and the back-propagation pass can then be expressed by
\begin{equation}
\delta_{x} = \sum_{i \in l} \frac{\partial \Pi_i}{\partial \mathbf{X}}.
\end{equation}
Please note that this framework is however not limited to a max-pooling subsampling operation but generalises to any pooling function.


\subsection{Multi-scale extraction}
\begin{figure}[!htbp]
\begin{center}
\includegraphics[width=.9\textwidth]{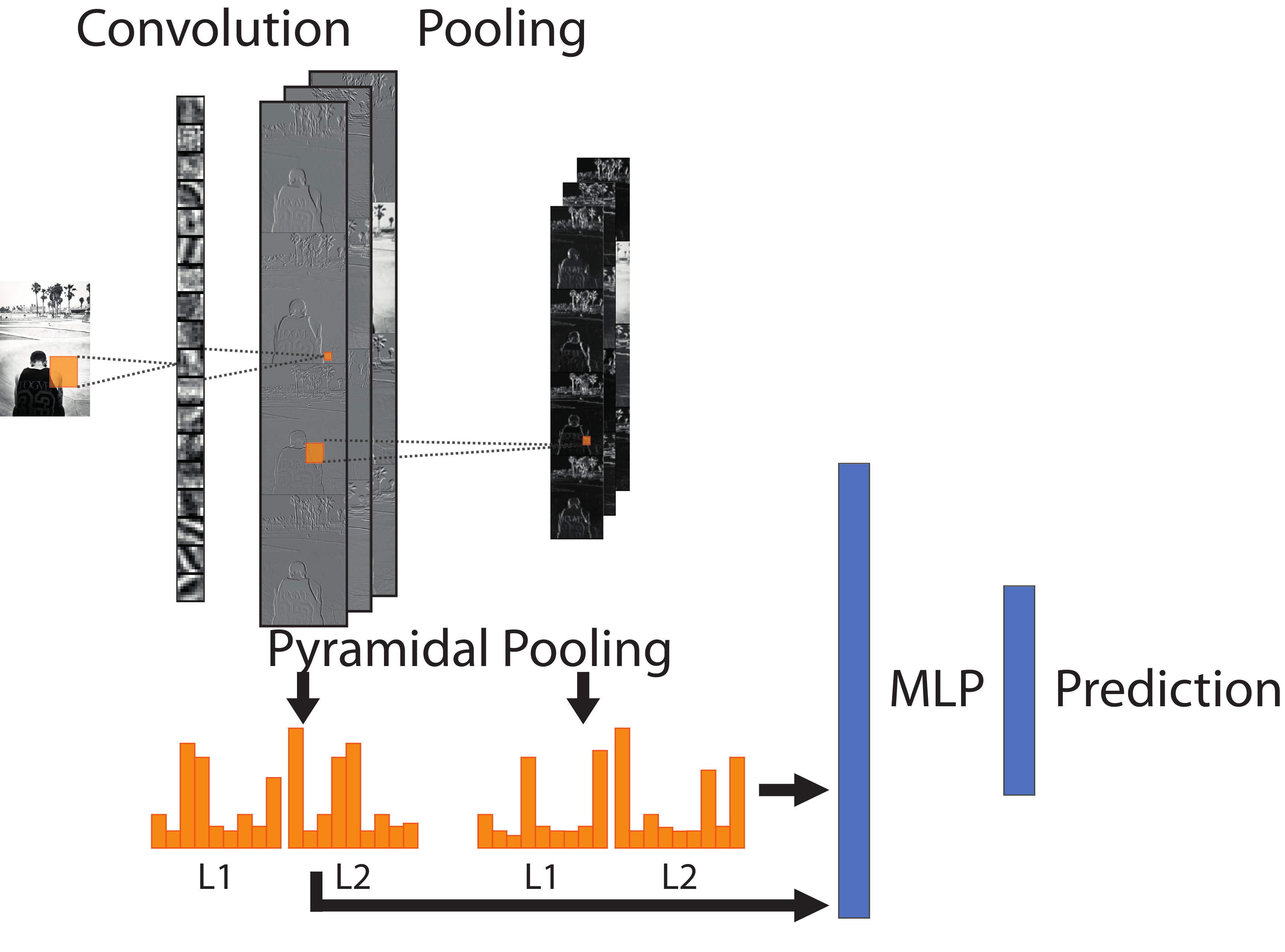}
\caption{Schematic representation of a Multi-Scale Pyramidal Pooling Network where histogram-like representations are extracted at two levels and at two scales. The first scale represents the output of the convolutional layer whereas the second scale is given by the output of a pooling (downsampling) layer. The resulting features are concatenated and used as input for a fully connected layer (MLP) which performs classification.}
\label{fig:pyrpool_net}
\end{center}
\end{figure}

A pyramidal feature extraction, while being itself already an improvement over the conventional CNN architecture as it allows to relax the constraint on a fixed input size, extracts features corresponding only to a single scale. We consider a scale the nominal size of the ``simplified'' image where the pooling is performed. In many applications, where input images come at very different scales, applying a pyramidal pooling layer will not completely solve the problem. Multi-scale pyramidal feature extraction can be done using a pyramidal pooling layer for each representation (i.e. layer in the network), and then concatenating the various feature vectors for the classification stage. After a subsampling layer the image gets down-sampled by a constant factor. Attaching a pyramidal pooling before and after a max-pooling operation therefore delivers a multi-scale extraction (Fig. \ref{fig:pyrpool_net}).

\subsection{Feature encoding layer}
The next step in narrowing the gap between the conventional bag-of-features approach and CNN is represented by introducing a feature quantisation layer, the major result of this paper.
Descriptors are generally quantised using an overcomplete dictionary resulting in an encoding with only a single or a few non-zero components.
This highly non-linear mapping seems to be crucial in obtaining good recognition rates in computer vision benchmarks. It seems then natural to implement an encoding layer also in our framework. If we consider the simplest of the feature quantisation algorithms, k-means with hard assignment, we note that such an encoding algorithm can be approximated by a correlation based measure 
\footnote{$||\mathbf{x}-\mathbf{y}||_2=\sqrt{\sum_i (x_i-y_i)^2}=\sqrt{\sum_i x_i^2 + \sum_i y_i^2 - 2\sum_i x_iy_i}$, and in the case of $\mathbf{x}$ and $\mathbf{y}$ normalised to have zero mean and unit variance reduces to the correlation between $\mathbf{x}$ and $\mathbf{y}$ as their sum will be almost constant.}. This allows us to derive a differentiable approximation of such coding scheme which we name as MLPDict.

\noindent \textbf{MLPDict Layer}: We use a fully connected layer with max-pooling to mimic the behaviour of the k-means coding. The projection of $\mathbf{x}$ with $\mathbf{W}$, the weights of the encoding layer, is the correlation between $\mathbf{x}$ and each column of $\mathbf{W}$. Taking the maximum correlation value is then an approximation of the hard assignment coding scheme. Compared to the bag-of-features approach, $\mathbf{W}$ serves as the dictionary, however an adaptive one which is tuned sample after sample. Moreover our encoding layer together with a non-linear activation function is powerful enough to reduce the dimensionality of the embedding producing much shorter feature vectors which can be processed very quickly in online systems.
Performing a pyramidal pooling operation on the result of such an encoding will produce a histogram-like representation.

\begin{figure}[htbp]
\begin{center}
\includegraphics[width=.9\textwidth]{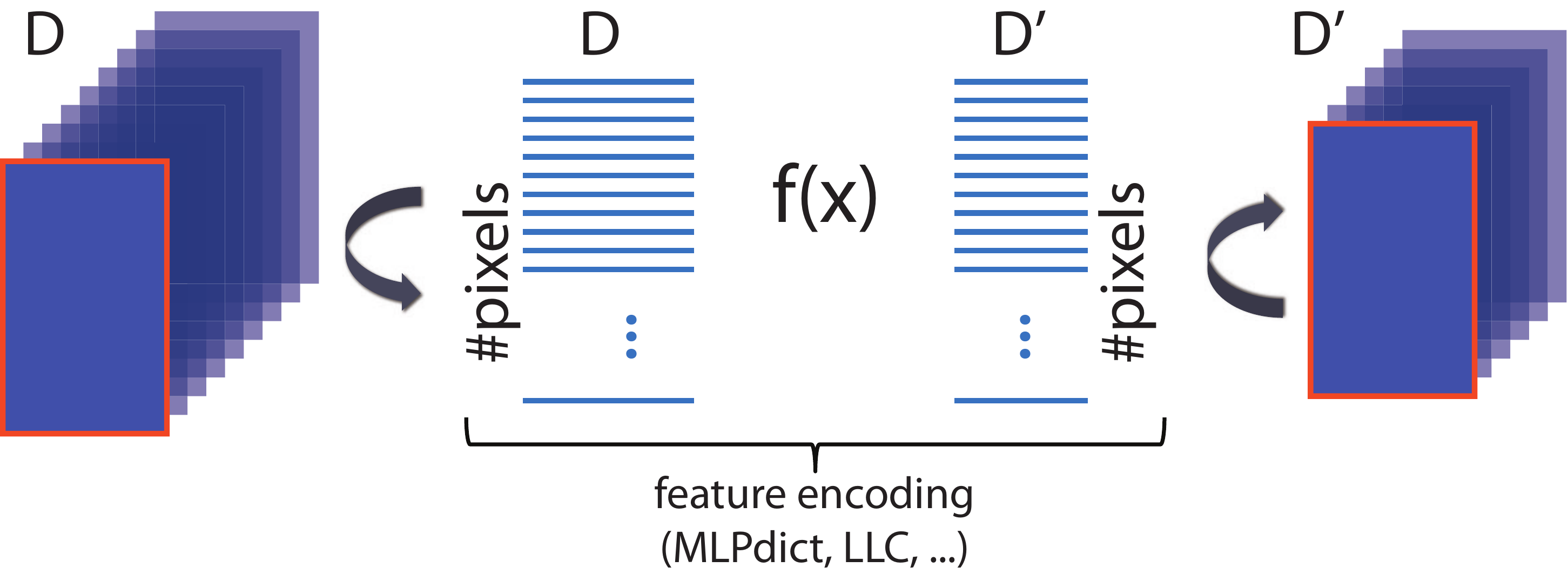}
\caption{The MLPdict layer used for feature encoding. Image responses of a network layer are reshaped to produce $D$ dimensional feature vectors, where $D$ represents the number of images in the layer. We consider each pixel as a descriptor and we map them onto another representation of size $D'$ for which only the maxima value per row is preserved.}
\label{fig:quantisation-layer}
\end{center}
\end{figure}

In Figure \ref{fig:quantisation-layer} a schematic representation of the encoding layer is shown. The hidden representation of a convolutional-based network is composed of $D$ images. We consider each pixel as a $D$ dimensional feature vector (extracted from the densest grid); for the encoding the set of images is reshaped into a matrix with as many rows as pixels in each image of the hidden states and as many columns as images $D$. Applying a fully connected MLP layer with a weight matrix $\mathbf{W} \in \mathbb{R}^{D \times D'}$  to the reshaped matrix $\mathbf{X} \in \mathbb{R}^{N \times D}$ will result in $\mathbf{X'} \in \mathbb{R}^{N \times D'}$ that is reshaped back onto $D'$ images, where $N$ corresponds to the number of pixels in each image. When $D' \ll D$ the layer acts as a feature selection layer which reduces redundancies in the responses and regularises the model when too many filters are used. This is common in image processing, especially in hyper-spectral data processing. When $D' \gg D$ the layer, thanks to the max-pooling operation, operates as a conventional encoding layer such as the one presented in section \ref{sec:featencoding}.
We refer to this quantisation layer as MLPdict and this is what we use in all our experiments.




\section{Experiments}
In all experiments, unless stated otherwise, we evaluate the average per-class accuracy to establish the classification performance, a more meaningful measure than overall recognition accuracy in case of unevenly distributed datasets. In all experiments no additional preprocessing, such as translation or deformation is used. All nets are trained using stochastic gradient descent with an initial learning rate of $0.001$, annealed by a factor of $0.97$ at every epoch and a momentum term of $0.9$.
We usually reached our best result in few training epochs, at most 30, whereas for CNN many more are in general required, especially when distortions and translations are added to the input. 
We validate our model on several benchmarks and we also show results of a challenging  application from the steel industry for which the MSPyrPool framework has been designed for. 

Comparing our architecture with CNN is not an easy task as the two approaches differ considerably. Nevertheless we try to make the comparison as fair as possible by taking equally sized convolutional and subsampling layers in both architectures and letting them differ for the choice of the encoding and classification stages.

\textbf{MNIST} As a reference benchmark to compare our approach with a conventional CNN we take the well studied MNIST \cite{lecun:1998} benchmark of handwritten characters. CNN excel on this dataset where all digits are of equal size and centred in the middle of a $28\times28$ grey-scale image. 
Only for this experiment we use the overall classification accuracy without averaging the per-class performance, to easily compare with other methods.
We use a MSPyrPool net with a convolutional layer with $5\times5$ filters and $100$ output maps, a $2\times2$ max-subsampling layer,
a pyramidal pooling with linear MLPDict at the output of the two subsampling layers with $l = \{1, 2, 4, 8\}$ and $l = \{1, 2, 4\}$ quadrants respectively. A single fully connected layer with softmax is used for classification.
Results are shown in Table \ref{tab:mnistResults}. It is interesting to note that our approach is the best among the fully supervised CNN approaches and on-par with the ones which use unsupervised pre-training while just using a very small single layer network and $100$ filters.

  \begin{table}[htdp]
\caption{Classification results for the MNIST benchmark. Our MSPyrPool network is compared with other CNN-based approaches which do not use any preprocessing of the input (e.g. no translations nor distortions).}
\begin{center}
\begin{tabular}{c | c c }
	 & Test \% \\
\hline
\hline
 \multicolumn{1}{r|}{ CNN LeNet-5 \cite{lecun:1998} } 	 & 99.05 \\
\hline
\multicolumn{1}{r|}{ CNN +  unsup. pre-training \cite{ranzato:2006nips}}  & 99.40 \\
\hline
\multicolumn{1}{r|}{ CNN +  unsup. pre-training \cite{masci:2011}}  & 99.29 \\
\hline
\hline
 \multicolumn{1}{r|}{ MSPyrPool }  & 99.13 \\
\hline
\end{tabular}
\end{center}
\label{tab:mnistResults}
\end{table}%
  
\textbf{CUReT}
The Columbia-Utrecht (CUReT) database \cite{curet} contains $61$ textures, and each texture has $205$ images obtained under different viewing and illumination conditions. Results are reported for all $61$ textures. For training our architecture only a single image is required as input, just as in previous work \cite{varma:2005}, with no information (implicit or explicit) about the illumination and viewing conditions.

We use the conventional evaluation protocol \cite{varma:2005} but with a different random split of the data. Images are normalised to have zero mean and unit variance to compensate for very different light conditions. We train a CNN with $5$ hidden layers: convolutional layer with $11\times11$ filters and $20$ output maps, $5\times5$ max-subsampling layer, convolutional layer with $9\times9$ filters and $20$ output maps, $5\times5$ max-subsampling layer, classification layer with softmax activation function. We use $tanh$ as activation function for every convolutional layer. We compare the result with a MSPyrPool net composed by: convolutional layer with $11\times11$ filters and $20$ output maps, $5\times5$ max-subsampling layer, convolutional layer with $9\times9$ filters and $20$ output maps, pyramidal pooling with MLPdict with $tanh$ activation and codebook size of $100$ at the output of the first and second convolutional layers. Features are pooled using $l = \{1, 2, 4\}$ and $l = \{1, 2, 3\}$ levels to produce a $3500$ dimensional vector.
An MLP with softmax activation performs the classification stage. Both networks minimise the multi-class cross-entropy loss commonly used in conjunction with a softmax activation. Training is stopped as soon as the training set is fully recognised.

In Table \ref{tab:curet} we compare results of our new MSPyrPool network with conventional CNN. The MSPyrPool net generalises much better to the unseen test data and demonstrates the superiority over conventional CNN for the task of texture classification. Furthermore we note that 98.7\% recognition rate on all $61$ classes is outperforming the bank-of-filter and texton based state-of-the-art 96.4\% \cite{varma:2005}, which share a similar architecture but with pre-wired feature extractors instead. This is a remark that the novel layers are a valuable contribution for the CNN framework and help generalisation.

\begin{table}[htdp]
\caption{Classification results for the CUReT benchmark. A conventional CNN is compared with our MSPyrPool network.}
\begin{center}
\begin{tabular}{c | c }
	 & Test \% \\
\hline
\hline
\multicolumn{1}{r|} { Bank-of-filters + Textons \cite{varma:2005} } & 96.4 \\
\hline
\hline
 \multicolumn{1}{r|}{ CNN } 		& 96.5 \\
\hline
 \multicolumn{1}{r|}{ MSPyrPool } 	& 98.7 \\
\hline
\end{tabular}
\end{center}
\label{tab:curet}
\end{table}%

\textbf{Caltech101} A further validation of the proposed system is performed on a classical pattern recognition benchmark, Caltech101, where CNN have never successfully been applied. 
We compare our results with those obtained with a similar system using the bag-of-feature paradigm \cite{Lazebnik:2006}. 
We use $30$ images per class for training and we test on at most $50$ of the remaining images converted to grey-scale. The feature encoding layer we adopted in this paper is in fact an approximation of the $k$-means quantisation with hard assignment. We consider a net with and without an encoding layer. Both nets are composed by: convolutional layer with $16\times16$ filters and $100$ output maps, $5\times5$ max-subsampling layer. For the net without an encoding layer a pyramidal pooling layer with $l=\{1,2,4\}$ is used to create a $2100$-dim feature vector that is classified by a fully connected net. For the net with an encoding layer, we used a dictionary size of $1024$ just before the pyramidal pooling layer. Using a pyramidal pooling layer with $l=\{1,2,4\}$ results in a $21504$-dim feature vector that is classified by a fully connected net. We train both a net with a linear and non-linear activation function in the encoding layer. For the sake of completeness we also train a CNN consisting of: convolutional layer with $16\times16$ filters and $100$ maps; $5\times5$ max-subsampling layer; convolutional layer with $13\times13$ filters and $100$ maps; $5\times5$ max-subsampling layer; fully connected classification layer. This CNN architecture is by no means the best architecture for this task, and is only listed to quantify the improvement using MSPyrPool nets. Results of all experiments together with results from the literature are listed in Table \ref{tab:caltech101}.

%
%

\begin{table}[htdp]
\caption{Classification results for the Caltech101 benchmark. A MSPyrPool net without an encoding layer (net1), with a linear (net2) and a nonlinear encoding layer (net3) are listed. }
\begin{center}
\begin{tabular}{c | c }
	 & Test \% \\
\hline
\hline
 \multicolumn{1}{r|}{ CNN }  & 25.2 \\
\hline
  \multicolumn{1}{r|}{ net1 }  & 52.8 \\
\hline
 \multicolumn{1}{r|}{ net2 }  & 57.9 \\
\hline
 \multicolumn{1}{r|}{ net3 }  & 55.2 \\
\hline
\hline
\multicolumn{1}{r|}{ Spatial Pyramid \cite{Lazebnik:2006}} & 64.6 \\
\hline
\multicolumn{1}{r|}{LLC \cite{wang:2010}}  & 73.4 \\

\end{tabular}
\end{center}
\label{tab:caltech101}
\end{table}%

MSPyrPool nets clearly improve recognition rate compared to a similar sized CNN.\footnote{Better results are obtained with deeper and bigger CNN, we got $40\%$ with a huge CNN still worse than MSPyrPool.} 
Using an encoding layer improves generalisation performance even though the resulting nets have many more free parameters. In this experiment using a non-linear activation degrades generalisation, probably due to the fact that the dictionary size is too big for the non-linear case, and better results might be obtained using much smaller dictionaries in the encoding layer. Nevertheless, the results show that we are able to jointly learn the feature extraction, the quantisation and the classification stages fully online.


\textbf{Steel-Defects}
Steel is a textured material and defects come at varying scales, which makes the task of classifying a wide range of defects extremely difficult. It is not easy to find a good resizing technique without destroying the original information content of the images. As a matter of fact if images/objects in a given classification task are varying over a few order of magnitudes it is even impossible to resize all the images. As a consequence one would need to split the training set and train various networks on subclasses of comparable size \cite{masci:2012ijcnn}. 
Using a MSPyrPool net avoids such problems because of the input-size independent feature extraction, one of the main contributions of this paper.

\begin{figure}[htbp]
\begin{center}
\includegraphics[width=.9\textwidth]{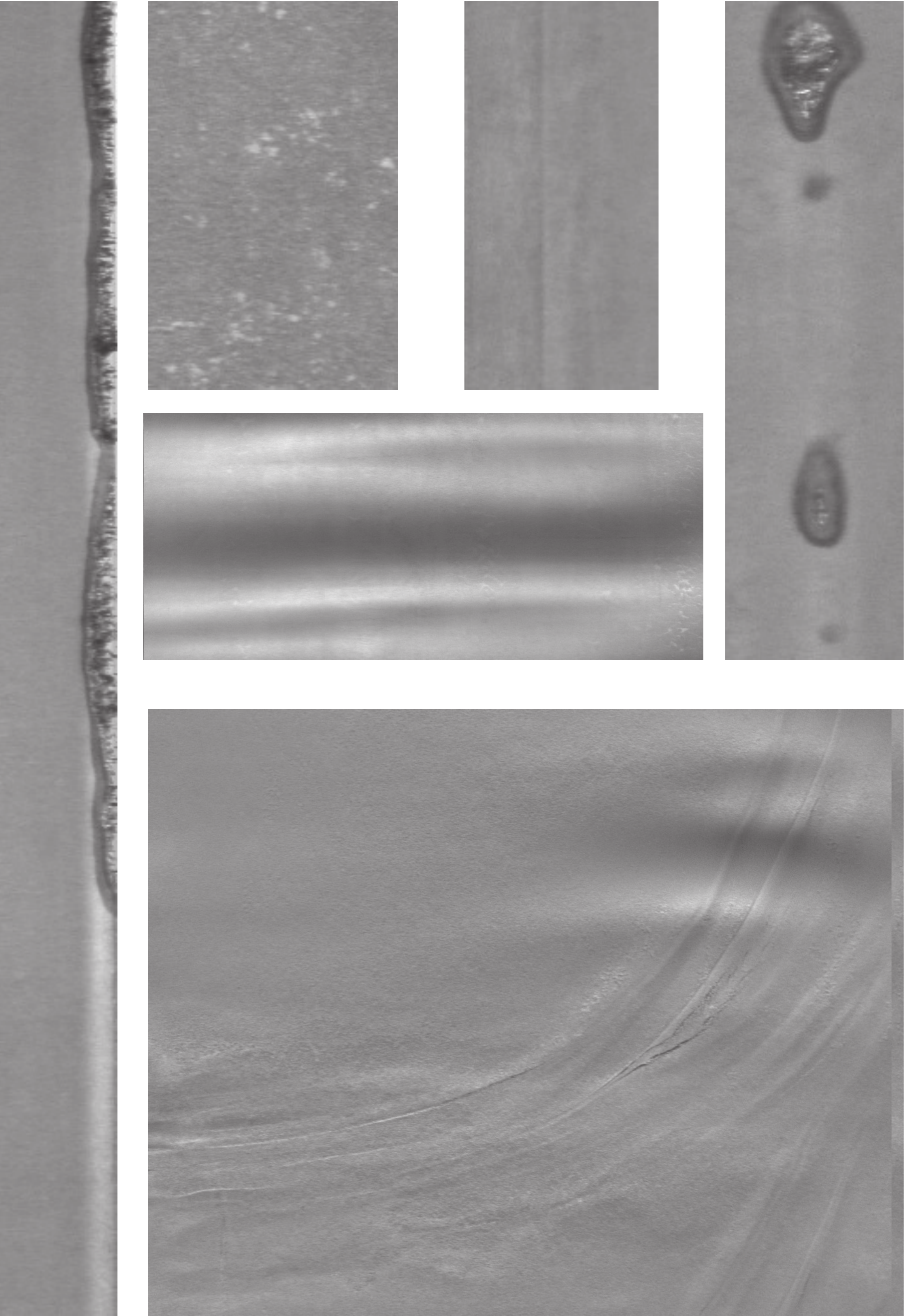}
\caption{ Subset of images from the Steel-defects benchmark showing the great difference in size among various samples. In this setting is not possible to resize the images to the same size as it will destroy most of them, hence a CNN is not applicable to solve this problem. }
\label{fig:dunk}
\end{center}
\end{figure}

For this experiment we use a proprietary dataset of ArcelorMittal from a hot-strip mill containing $34$ different defect classes. A region-of-interest (ROI) is provided for each of the instances which vary greatly in size from a minimum edge length of $\approx 20$ to a maximum of $\approx 2000$ pixels (see Fig. \ref{fig:dunk} for a subset of defects). Furthermore the dataset is unevenly distributed w.r.t. the number of samples per class. To obtain a good support to perform the pyramidal pooling we add background information to get a minimum patch size of $100$ pixels along each dimension. We also limit the maximum size per dimension to $500$ pixels to accelerate training by resizing images to have at most $500$ pixels for the longest edge.


We compare our system to a set of classifiers trained on commonly used features such as (LBP, HOG, PHOG) using the same training setup described in \cite{masci:2012ijcnn} but with a deeper MLP for classification ($500$ and $250$ units). 
We use a MSPyrPool with: convolutional layer with $11\times11$ filters and $20$ output maps, $2\times2$ max-subsampling layer, convolutional layer with $9\times9$ filters and $20$ output maps, classification layer with softmax activation function. $tanh$ is used for every convolutional layer. Pyramidal pooling layers with MLPdict, $tanh$ activation and codebook size of $100$ are attached at the output of the first and second convolutional layers. Features are pooled at $l = \{1, 2, 4\}$ and $l = \{1, 2, 3\}$ levels producing a vector of $3500$ dimensions.

The results are summarised in Table \ref{tab:dunk}. 
We see that a MSPyrPool net can be applied to solve a problem where conventional CNN can not be applied. In this setting resizing will basically destroy the images and will remove the actual defect size information. We also see that the performance of our model, which learns everything from pixel representation with no prior knowledge, outperforms any of the single features classifiers.

\begin{table}[htdp]
\caption{Classification results for the Steel-defect benchmark where CNN fail. Various classifiers trained on conventional features (LBP, HOG, PHOG)
 are compared with our MSPyrPool network. Our method outperforms any classifier based on classical computer vision features.}
\begin{center}
\begin{tabular}{c | c c}
							 & Test \% \\
\hline
\hline
\multicolumn{1}{r|}{ LBP }       	    	& 29.1 \\
\multicolumn{1}{r|}{ LBP-HF } 	    	& 48.9 \\
 \multicolumn{1}{r|}{ MONO-LBP } 	& 64.0 \\
 \multicolumn{1}{r|}{ VAR } 		& 34.3 \\
 \multicolumn{1}{r|}{ HOG } 	   	& 60.5 \\
 \multicolumn{1}{r|}{ PHOG } 	   	& 58.9 \\
\hline
\hline
 \multicolumn{1}{r|}{ CNN } 	& - \\
 \hline
 \hline
 \multicolumn{1}{r|}{ MSPyrPool } 	& 67.9 \\
\hline
\end{tabular}
\end{center}
\label{tab:dunk}
\end{table}%


\section{Conclusions}
We proposed an extension of conventional CNN consisting of three new ingredients: 1) a pyramidal pooling layer that makes the net independent of input image size; 2) a multi-scale feature extraction;  3) an encoding layer emulating standard dictionary-based encoding strategies. 
We validated the novel architecture on various benchmarks and obtained results comparable to or better than the current state-of-the-art. The full potential becomes evident when images for a given classification task vary in size, applications that so far have eluded CNN. The proposed extensions open up the possibility to use fully supervised convolutional-based neural nets in many new applications. 

On the other hand, this work closes the gap between neural network-based models and weak classifiers such as the bag-of-features approach. To the best of our knowledge, ours is the first architecture that learns feature extraction and encoding online and in fully supervised fashion.

We believe that more complex encoding layers may considerably
boost recognition rates of our Multi-Scale Pyramidal Pooling network,
a new and flexible framework for supervised object classification.

%
%
\bibliographystyle{splncs}
\bibliography{references}

\end{document}